%% file: main.tex
\pdfoutput=1

\documentclass[11pt]{article}

\usepackage[]{EMNLP2024}

\usepackage{times}
\usepackage{latexsym}

\usepackage[T1]{fontenc}

\usepackage[utf8]{inputenc}

\usepackage{microtype}

\usepackage{inconsolata}

\usepackage{graphicx}
\usepackage{amsmath}
\usepackage{multirow}
\usepackage{subcaption}

\usepackage{booktabs}
\usepackage{xspace}
\usepackage{makecell}

\usepackage{siunitx} %

\usepackage[capitalize,noabbrev]{cleveref}

\usepackage{anyfontsize}

\newcommand{\llama}{Llama\xspace}

\def\tok100{\mathcal{T}_{100}}

\input{macros/macros}

\input{macros/std}
\input{macros/math_commands}

\usepackage{xcolor}

\usepackage{fdsymbol}

\title{Target-Aware Language Modeling via Granular Data Sampling}

\author{
Ernie Chang$^{\spadesuit}$ \,
Pin-Jie Lin$^{\clubsuit}$ \,
Yang Li$^{\vardiamondsuit}$ \,
{\bf Changsheng Zhao$^{\spadesuit}$} \,\\
{\bf Daeil Kim$^{\spadesuit}$} \,
{\bf Rastislav Rabatin$^{\spadesuit}$} \,
{\bf Zechun Liu$^{\spadesuit}$} \,
{\bf Yangyang Shi$^{\spadesuit}$} \,
{\bf Vikas Chandra$^{\spadesuit}$} \\
$^\spadesuit$AI at Meta \\
$^\clubsuit$Virginia Tech \\
$^\vardiamondsuit$Iowa State University \\
{\tt erniecyc@meta.com, pinjie@vt.edu, yangli1@iastate.edu}
}

\begin{document}

\maketitle

\begin{abstract}

\input{sections/00_abstract}

\end{abstract}
\input{sections/01_introduction}

\input{sections/02_related}
\input{sections/03_approach}

\input{sections/04_experiments}

\input{sections/05_results}

\input{sections/06_conclusion}

\bibliography{anthology,bibtex/coreset,bibtex/custom,bibtex/datasets,bibtex/nas-overview,bibtex/pretrained,bibtex/zero_nas,bibtex/mllm,bibtex/sampling}
\bibliographystyle{acl_natbib}

\clearpage

\input{sections/appendix}

\end{document}

%% file: macros/macros.tex
\newcommand{\qfeat}{q_{\text{feat}}}
\newcommand{\pfeat}{p_{\text{feat}}}

\DeclareMathOperator*{\argmin}{arg\,min}

\newlength{\widebarargwidth}
\newlength{\widebarargheight}
\newlength{\widebarargdepth}

%% file: macros/std.tex
\newcommand\sX{\ensuremath{\mathcal{X}}}

\newcommand\sZ{\ensuremath{\mathcal{Z}}}

\newcommand{\1}{\mathbb{I}} %

%% file: macros/math_commands.tex
\usepackage{amsmath,amsfonts,bm}

\def\eqref#1{equation~\ref{#1}}

\def\1{\bm{1}}

\DeclareMathAlphabet{\mathsfit}{\encodingdefault}{\sfdefault}{m}{sl}
\SetMathAlphabet{\mathsfit}{bold}{\encodingdefault}{\sfdefault}{bx}{n}

\def\sX{{\mathbb{X}}}

\def\sZ{{\mathbb{Z}}}

%% file: sections/00_abstract.tex
Language model pretraining generally targets a broad range of use cases and incorporates data from diverse sources. 
However, there are instances where we desire a model that excels in specific areas without markedly compromising performance in other areas. 
A cost-effective and straightforward approach is sampling with low-dimensional data features, which allows to select large-scale pretraining data for domain-specific use cases.
In this work, we revisit importance sampling with n-gram features consisting of multi-granular tokens, which strikes a good balance between sentence compression and representation capabilities.
We observed the sampled data to have a high correlation with the target downstream task performance \emph{while preserving its effectiveness on other tasks}. 
This leads to the proposed data sampling paradigm where language models can be pretrained more efficiently on selected documents.
On eight benchmarks we demonstrate with $\sim$1\% of the data, pretrained models perform on par with the full RefinedWeb data and outperform randomly selected samples for model sizes ranging from 125M to 1.5B.

%% file: sections/01_introduction.tex
\section{Introduction}\label{section:introduction}
Language model pretraining is the cornerstone of universal language models (LMs), creating general-purpose representations to excel across a variety of NLP downstream tasks~\citep{doi:10.1080/00401706.1975.10489266,10.5555/2380985}.
This process often involves the use of vast amounts of text, sometimes measured in billions or even trillions of tokens from webpages \cite{abnar2022exploring,kaplan2020scaling}.
However, there are instances where a model needs to perform well in specific domains while not compromising performance in others. 
This necessitates the use of data selection methods to determine which potential data points should be included in the training dataset and how to effectively sample from these selected points~\citep{albalak2024survey}.

One approach to reducing data size is \emph{coreset selection}, which involves selecting a small, representative subset of data~\citep{pmlr-v162-du22c}. 
Coresets can significantly decrease computational costs while maintaining robust performance. 
In this work, we explore the optimization of coresets towards a target data distribution, but relaxing the data sampling process to reduce domain biases.

\input{grid_figs/overview}
Here we revisit importance sampling~\cite{rubin1988resampling,xie2023data} by proposing to utilize tokens of different granularities as features, ranging from subword, word, to multi-word (or n-gram) tokens (Shown in Figure~\ref{fig:overview}). 
We observed empirically that by controlling the granularity of tokens in the tokenizer, we can construct coresets with less domain biases -- fine-grained tokens capture introduces more task knowledge while coarse-grained tokens preserve general information.
Thus, we experiment with adapting the vocabulary set of pretrained tokenizers to data from specific tasks, and modulating token granularity in the vocabulary to maintain generality.
To demonstrate the efficacy of multi-granular sampling, we use eight downstream tasks as target tasks with Llama-3's tokenizer being the base tokenizer, where its vocabulary set is adapted to task data. 
This process creates a domain-specific vocabulary set which we use to featurize text documents\footnote{We employ n-gram features in the same way as \citet{xie2023data}.} for target-aware data sampling with higher quality sampled data. 
Our work demonstrates that smaller language models pretrained on the coresets perform well across four various model sizes ranging from 125M to 1.5B. 
This contributes to the ongoing discourse on optimizing pretraining data for improved computational efficiency and model performance.
Unlike past approaches which can be computationally-intensive~\cite{wenzek-etal-2020-ccnet,wettig2024qurating,muennighoff2024scaling} or simple but easily biased towards target data distribution~\cite{xie2023data}, our approach is unique in two significant ways:
\begin{enumerate}
\item We proposed an algorithm to merge a pretrained tokenizer with multi-granular tokens and empirically showed that it yields highly efficient n-gram features that has high correlation with downstream task performances.
\item Leveraging our findings, we improve upon the importance-based data sampling technique by adapting a general vocabulary set to the target vocabulary.
This creates a better representation of data that enhances model performances in target tasks, while maintaining decent performance in non-target tasks.
\end{enumerate}

%% file: grid_figs/overview.tex
\begin{figure}[t]
  \centering
\includegraphics[width=1.1\columnwidth]{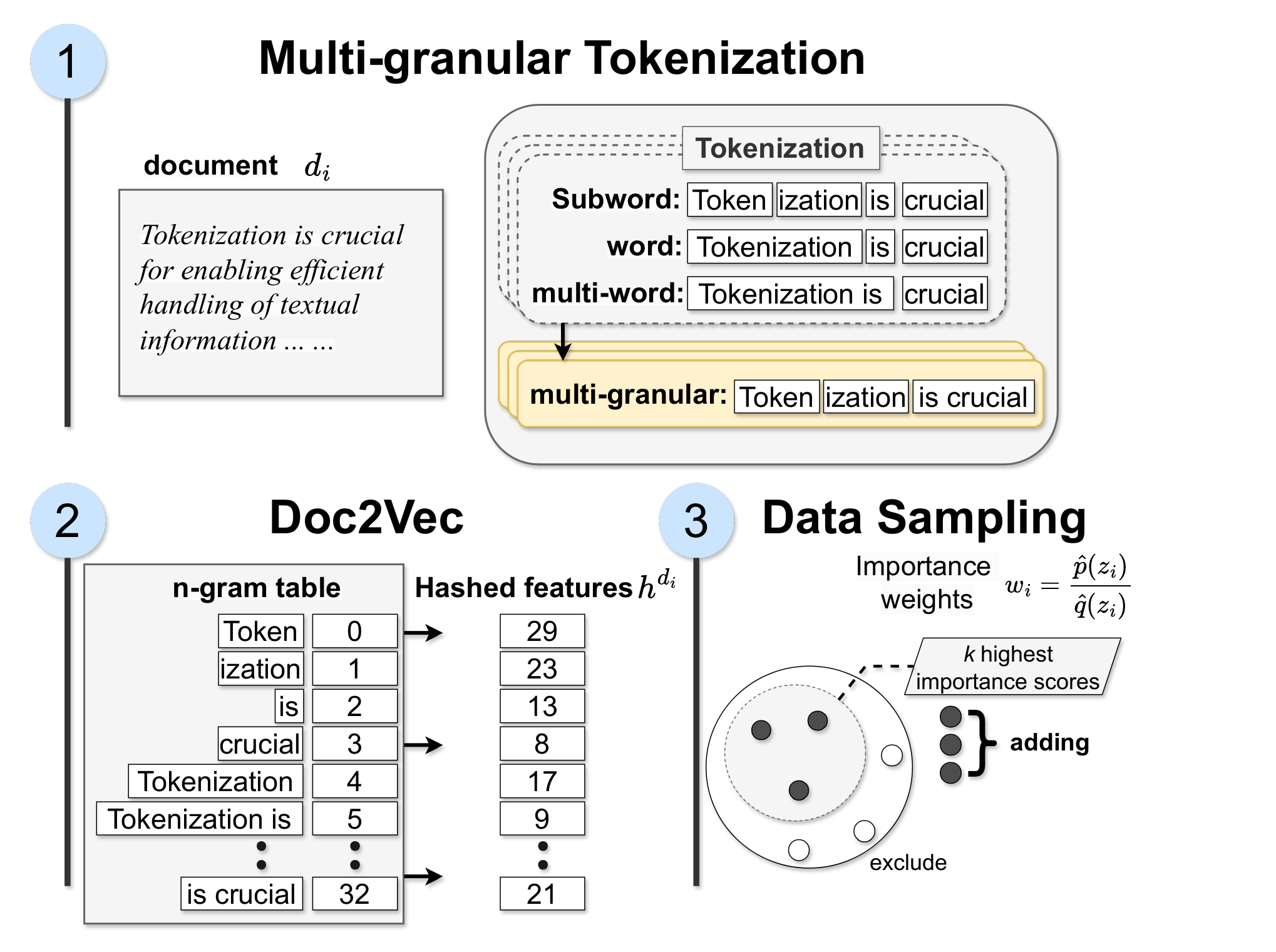}
\caption{\small Multi-granular tokenization for more modular feature vectors used in importance sampling.
(1) Given a document $d_{i}$, it undergoes featurization as a sequence of multi-granular tokens. (2) Subsequently, the document is transformed into a fixed-sized feature representation via hashing N-grams. (3) We measure its significance through the enhancement weight $w_i$ and select a subset of $K$ representative data points from the original target distributions through re-sampling}
\label{fig:overview}
\end{figure}

%% file: sections/03_approach.tex
\section{The Approach}\label{section:methodology}

Selecting samples from large-scale datasets such as RefinedWeb~\cite{penedo2023refinedweb} is slow and expensive.
A tractable solution is to encode each document as a vector using n-gram features that can be computed easily. 
Here we assume in our settings a small number of target text examples $D_{task}$ from a target distribution $p$ and a large raw dataset $D_{raw}$ drawn from a distribution $q$ with $N$ examples, we aim to select $k$ examples ($k \ll N$) from the raw dataset that are similar to the target.

We adopted the importance sampling technique as in \citet{xie2023data} which selects examples that align with target distribution. 
The technique provides a tractable importance estimate of each text, and applies 
importance sampling on a feature space $\sZ$ that provides the necessary structure. 
The feature extractor $h: \sX \rightarrow \sZ$ is used to transform the input $x$ into features $z=h(x)$. 
The resulting raw and target feature distributions are $\qfeat$ and $\pfeat$, respectively. 
Our objective is to select examples whose features align with the target feature distribution $\pfeat$. 
To do so, features $\qfeat$ and $\pfeat$ are extracted (Figure~\ref{fig:overview}) using n-grams extracted from each tokenized document using an adapted tokenizer.
Each n-gram is mapped to a key in the hash table where the ids of the table define a fixed-size embedding, and each key maps to the n-gram count. 
Then, the importance weights is computed for each featurized example $z_i = h(x_i)$ from the $N$ raw examples, with the weight $w_i = \frac{\hat{p}_{\text{feat}}(z_i)}{\hat{q}_{\text{feat}}(z_i)}$. 
The final step involves sampling, where we select $k$ examples without replacement from a categorical distribution, the probabilities of which are given by $\frac{w_i}{\sum_{i=1}^N w_i}$. 
\input{tbls/0_main_result}

\paragraph{Tokenizer  Adaptation.} 
Here we adapt the vocabulary to the target data. 
To derive target vocabulary $V(t)$, we use Llama-3 tokenizer's vocabulary $V_{start}$ as the starting point and merge $V_{start}$ with $V_{task}$ which is learned from task data $D_{task}$.
In constructing $V_{task}$, we make sure to include multi-granular tokens (i.e. words and multi-words), where $V_{task}$ is then merged with $V_{start}$ to form $v(t - 1)$.
Next, we incrementally remove tokens from $v(t - 1)$ to obtain $v(t)$, where we minimize the distance from the original vocabulary set such that a less biased document feature can be extracted as n-gram vectors.
We first define a metric to measure the quality of vocabulary set on a corpus, following~\citet{xu2021vocabulary}, which proposed to learn optimal vocabulary by maximizing the vocabulary utility metric ($\mathcal{H}_{v}$)
computed as:

{\small
\begin{equation}
    \mathcal{H}_{v} = - \frac{1}{l_{v}}\sum_{j \in v } P(j)\log P(j), %
\end{equation}}
\input{grid_figs/0_vocab_size_vs_params}

where $P(j)$ is the relative frequency of token $j$ from the target data and $l_{v}$ is the average length of tokens in vocabulary $v$. 
For any vocabulary, its entropy score $\mathcal{H}_{v}$ can be calculated based on a vocabulary from its previous step. 
The optimization problem can be formulated as:

{\small
\begin{equation}
 \argmin_{v(t-1), v(t)}  \big [ \mathcal{H}{v(t)} - \mathcal{H}{v(t-1)} \big ]    
\end{equation}
}
where $v(t)$ and $v(t - 1)$ are two sets containing all vocabularies with upper bound of size $|v(t)|$ and $|v(t - 1)|$ respectively.
In our implementations, we set $|v(t)| = 10k$, where $t=10$; and $|v(0)|$ is the default Llama-3 tokenizer's vocabulary size.
Here $\argmin$ aims to find the vocabulary from ${V}_{start}$ with the minimum entropy difference. 
In this optimization process, we include varying granularity of tokens in the vocabulary ranging from n-gram to multi-granular tokens. 
To evaluate the effectiveness of this approach, we compare the overall KL reduction and downstream task performances with different granularities. 
We observed that a mix of all granularities yields the best results overall on the downstream task (See Figure~\ref{fig:correlation}) --
where there is a clear trend of increasing performance with mixed granularities. 
However, a finer granularity also decreases the representation power of the features, as seen from the degradation in using subword tokens alone.
Empirically, we found that the proposed tokenizer adaptation technique yields significant advantage over naive merging of two vocabulary sets (See Appendix).

%% file: tbls/0_main_result.tex
\begin{table*}[t]
    \centering
\resizebox{0.9\textwidth}{!}{
\begin{tabular}{c c c c c c c c c c}
\toprule
\multirow{2}{*}{\textsc{Approach}} & \multicolumn{9}{c}{\textsc{}} \\ 
& \textsc{ARC-Easy} & \textsc{ARC-Hard} & \textsc{BOOLQ} & \textsc{PIQA} & \textsc{SIQA}  & \textsc{HELLASWAG} & \textsc{OBQA} & \textsc{WINOGRANDE} & \textsc{Avg.} \\ 
\midrule
\multicolumn{9}{c}{125M params} \\ \midrule
\textsc{random} & 45.74 & 27.64 & 59.38 & 66.41 & 41.02 & 37.13 & 34.77 & 52.64 & 45.59
 \\
\textsc{n-gram} & 43.59 & 27.20 & 57.76 & 72.17 & 42.29 & 45.89 & 31.05 & 50.24 & 46.27 \\
\textsc{Multi-granular} & 44.24 & 30.13 & 57.98 & 72.71 & 41.16 & 46.68 & 35.74 & 52.10 & \textbf{47.59} \\
\midrule
\multicolumn{9}{c}{350M params} \\ \midrule
\textsc{random} & 52.12 & 29.20 & 62.38 & 69.04 & 42.92 & 46.21 & 40.23 & 54.39 & 49.56 \\
\textsc{n-gram} & 49.58 & 30.52 & 62.23 & 75.68 & 42.38 & 57.18 & 40.72 & 53.91 & 51.52 \\
\textsc{Multi-granular} & 49.28 & 31.74 & 60.06 & 76.51 & 42.09 & 56.36 & 41.99 & 53.88 & \textbf{51.61} \\
\midrule
\multicolumn{9}{c}{500M params} \\ \midrule
\textsc{random} & 54.17 & 31.84 & 59.86 & 70.85 & 43.07 & 49.02 & 39.55 & 56.49 & 50.61 \\
\textsc{n-gram} & 49.65 & 31.02 & 63.09 & 76.29 & 42.83 & 58.17 & 41.56 & 54.47 & 52.13 \\
\textsc{Multi-granular} & 52.64 & 30.71 & 53.86 & 76.56 & 43.02 & 60.40 & 47.59 & 54.59 & \textbf{52.42} \\
\midrule
\multicolumn{9}{c}{1.5B params} \\ \midrule
\textsc{random} & 58.89 & 32.23 & 51.56 & 72.07 & 42.58 & 55.05 & 41.80 & 57.71 & 51.49 \\
\textsc{n-gram} & 53.91 & 34.28 & 60.57 & 79.49 & 44.48 & 66.42 & 40.43 & 54.49 & 54.26 \\
\textsc{Multi-granular} & 55.47 & 34.28 & 59.11 & 78.22 & 43.02 & 69.45 & 39.84 & 56.88 & \textbf{54.53} \\
\bottomrule
\end{tabular}}
    \caption{\small Results over all downstream tasks selecting based on all task validation sets, each in terms of its respective metric. Here we compare \textbf{Random}, \textbf{N-gram}, and \textbf{Multi-granular} data selection techniques sampling for 1\% data of RefinedWeb~\cite{penedo2023refinedweb} and pretrained with $\sim$700 million tokens. We observe two major trends: (1) performance improves with increase in number of parameters. However, the improvement begins to plateau as model becomes larger. (2) \textbf{Multi-granular} has the best overall performance across all benchmarks, despite being worse on some individual tasks.}
    \label{tab:main_results}
\end{table*}

%% file: grid_figs/0_vocab_size_vs_params.tex
\begin{figure}[t]
\centering
\includegraphics[width=\linewidth]{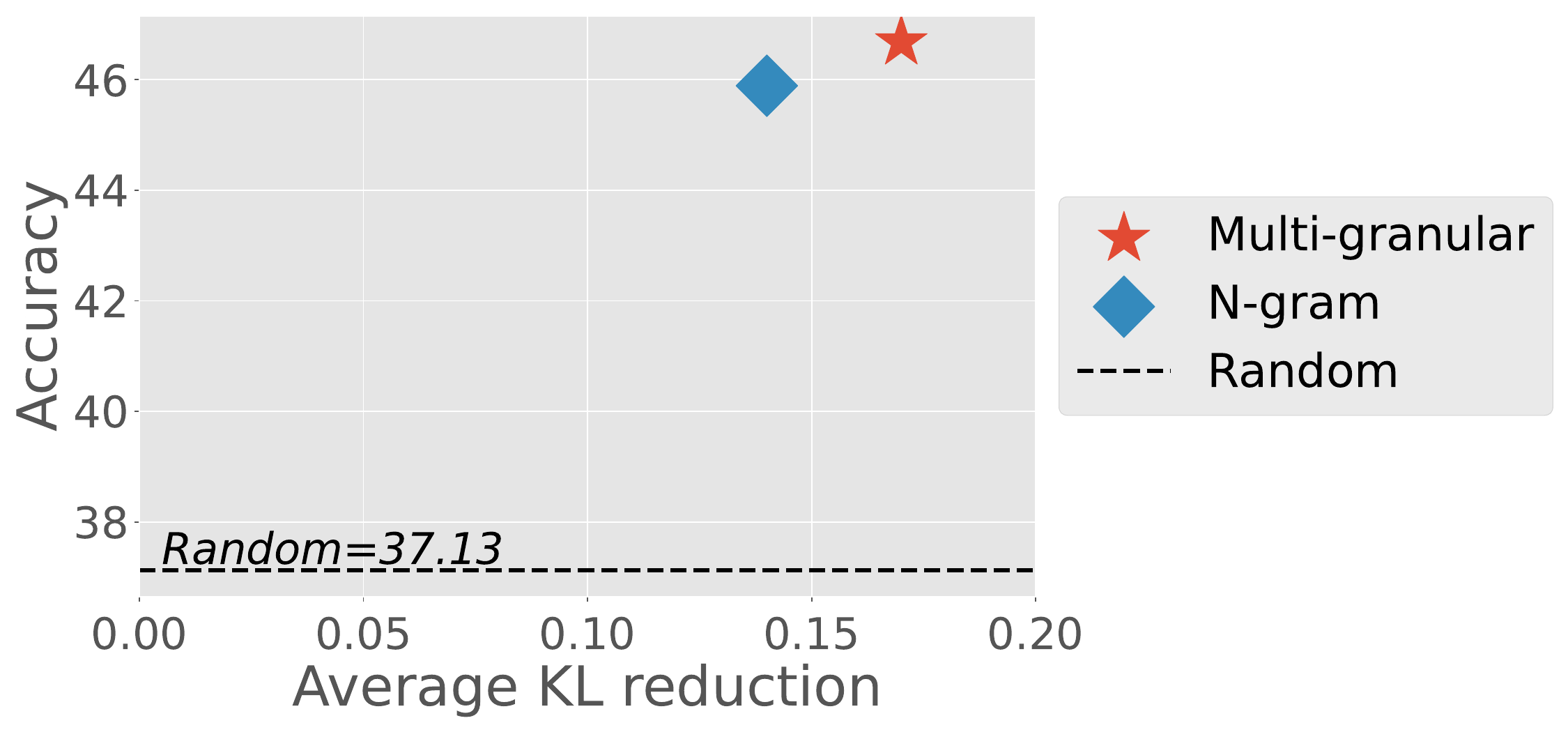}
\caption{\small
The plot of average KL reduction and the performance on HellaSwag. We measure how the granularity of tokens used used for coreset selection reduces KL divergence to the target distribution compared to random sampling from The RefinedWeb, suggesting a strong correlation between KL reduction and downstream performance (Pearson $r=0.82$).}
\label{fig:correlation}
\end{figure}

%% file: sections/04_experiments.tex
\section{Experimental Setup}
\textbf{Network, training details and evaluation.} 
We pretrain the decoder-only transformer using causal language modeling objectives on selected datasets, averaging over three initialization runs for each configuration, where model weights were randomly initialized.
The language models varied in size, with 125M, 350M, 500M, and 1.5B parameters. 
This range allowed us to explore how model complexity impacts the final results. 
Pretraining was conducted on a distributed computing setup with 32 GPUs across 4 nodes, each equipped with an H100 graphics card.
We evaluated our proposed \textbf{Multi-granular} selection approach against random selection (\textbf{Random}) and compared it with the same sampling algorithm using word-based \textbf{N-gram} features. Importance sampling~\cite{xie2023data} was employed for all feature types.

\paragraph{Datasets.} 
We evaluate the models on eight common sense reasoning tasks in a zero-shot fashion, including ARC-easy, ARC-challenge~\cite{clark2018arc}, BoolQ~\cite{clark2019boolq}, PIQA~\cite{bisk2020piqa}, SIQA~\cite{sap2019siqa}, HellaSwag~\cite{zellers2019hellaswag}, OBQA~\cite{mihaylov2018obqa}, and WinoGrande~\cite{sakaguchi2021winogrande}.

%% file: sections/05_results.tex
\input{grid_figs/2_emergent}

\section{Main Results}

Overall, language models varied in four sizes display marked improvements when trained on sampled coresets selected using multi-granular features, achieving a $6.94$\% improvement in average benchmark scores ( \textsc{Avg.}). 
The proposed method with multi-granularity consistently outperforms importance sampling that uses only n-gram features on the average of all benchmarks. 
Moreover, our result also shows that despite sampling based upon a target dataset, the model performance does not degrade on non-target benchmarks (See Figure~\ref{fig:domain}). 
Figure~\ref{fig:emergent} shows the sharp metric improvements of averaged performance on the eight tasks, starting at a model of size 125M to 1.5B, which indicates the potential of the technique to scale up the capabilities of small language models.

\section{Further Discussion}

\paragraph{Finer-grained Features Reduces Task Biases.}
Based on our ablation, we observe marked improvement by simply using subword n-grams. 
Moreover, we show in Figure~\ref{fig:domain} that selecting from a single task introduces task data biases that degrades the performance.
This is mitigated through the use of finer-grained n-gram features where we introduce multi-granular tokens containing subwords and multi-words, which gives an additional 5.78\% improvement over word-based n-grams. 
We postulate that this improvement has to do with the reduction of hash collisions in the hashed n-gram features, where the joint use of subword and multi-word capture beyond the boundaries of a word while preserving parts of a word in tokens so that during collisions the whole word is not entirely discarded and not represented (More analysis in Appendix~\ref{appendix:merger_1}). 

\input{grid_figs/3_domain_comparisons}

\paragraph{Impact of Multi-Granularity.}
We ablate the percentages of subword, word, and multi-word tokens on a subset of RefinedWeb.
Subword, word, and multi-word tokens are bounded by the vocabulary size, and generally, we found that multi-granular outperforms single-granular tokens.
This amounts to 3.23\% margin on zero-shot performance for a 125M model. 
Further, we run the experiments to vary the distribution of all three granularities at a fixed vocabulary size; 
where we found that a higher percentage of subword (60\%) to be the most viable, along with mixing with some word (30\%) and multi-word (10\%) tokens. 
Subword tokens capture more finer-grained details so they are more representative, however, they also make the sampling process slow as they make the sequence length longer.
Word and multi-word compresses sequence length so they reduce the impact of subword has on the sampling speed.

%% file: grid_figs/2_emergent.tex
\begin{figure}[t]
  \centering
\includegraphics[width=0.95\columnwidth]{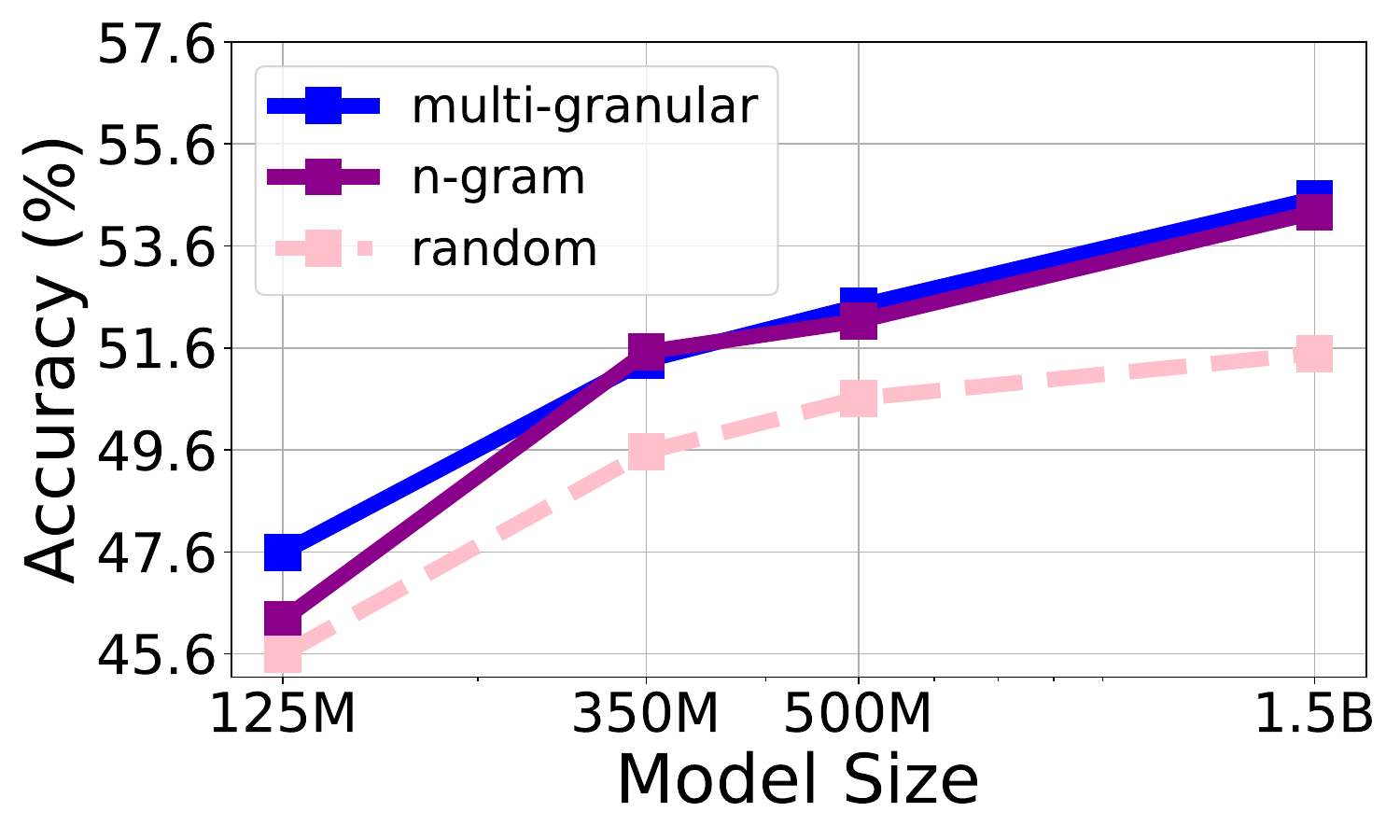}
\caption{\small Zero-shot performances averaged over eight tasks computed across all model sizes, where emergent characteristic can be observed at the model of size 350M parameters.}
\label{fig:emergent}
\end{figure}

%% file: grid_figs/3_domain_comparisons.tex
\begin{figure}[t]
\centering    
\includegraphics[width=\columnwidth]{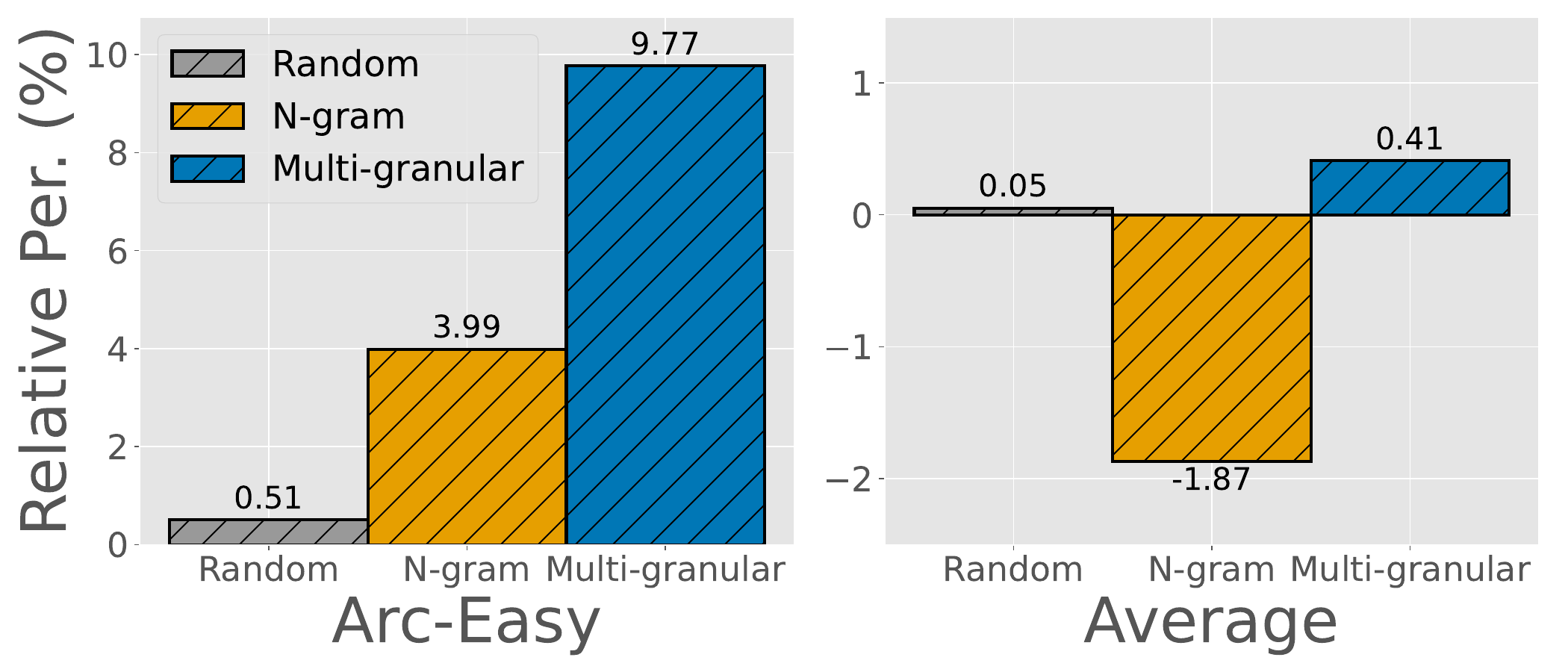}
    \caption{\small 
Comparison of \textbf{Multi-granular} n-grams with \textbf{N-gram} and \textbf{Random} baseline across eight tasks using 125M models, trained solely on data selected with ARC-Easy data as the target. Relative performance is used. We observe that multi-granular features enable the model to consistently outperform the baseline despite task-specific biases in the data. 
We performed similar experiments for all other benchmarks in the appendix, where for all eight tasks, the same pattern is observed that multi-granular n-grams yield almost no degradation across benchmarks.}
\label{fig:domain}
\end{figure}
\vspace{-5pt}

%% file: sections/06_conclusion.tex
\section{Conclusion and Future Works}\label{section:conclusion}
In this study, we revisited importance sampling of text corpus in language modeling by exploring multi-granular n-grams as features.
This led us to explore a pretraining paradigm where we can obtain more targeted data for more efficient language modeling.
Our findings, validated across eight benchmarks, allow us to put forward multi-granular n-grams features as viable document representations used in importance sampling. 
For future work, we will aim to extend this approach to larger language models and datasets.

\section*{Limitations}

While the method of targeted data sampling using low-dimensional features is efficient and enhances specific performance areas, it is not without its challenges. 
Further exploration is needed to refine the process of selecting optimal data features that balance domain specificity with general applicability. 
Importantly, we have not taken explicit steps to ensure that the sampled data does not contain biases in the data.
Moreover, we believe a more solid conclusions can be drawn when an even larger pretraining data is experimented, and other model-based approaches are also taken into account.
All in all, this study highlights the importance of fine-tuning data selection in pretraining smaller language models to avoid overfitting while maintaining robust performance across diverse tasks.

\section*{Ethics Statement}

The practice of selective data sampling in language model pretraining has shown promising results in enhancing model performance in targeted tasks. Our experiments are conducted using datasets that are widely recognized and utilized within the research community, ensuring the reproducibility and reliability of our results. However, the application of this method to sensitive or private datasets necessitates stringent adherence to ethical standards. Furthermore, the increased efficiency in training specialized smaller models could potentially lead to escalated computational demands, which must be considered when scaling these methods.

%% file: sections/appendix.tex
\onecolumn

\appendix

\section{Appendix}\label{appendix:details}

\subsection{Details on Document Sampling}\label{appendix:merger_1}

Here we provide additional details regarding the process of feature extraction from documents.
Due to the memory constraints on the machines, we split the RefinedWeb data into 16 shards, and sampled a subset from each shard based on the target data. 
This process takes around 1.5 days on average for all approaches, meaning that the change in tokenizer's vocabulary does not result in noticeable differences in sampling speed, since vocabulary also defines sentence compression ratio.

\paragraph{Analysis of Sampled Data.}

Further, we analyze the sampled data using various tokenization techniques. 
Here we provide the statistics over each technique in Table~\ref{tab:data_stats}. 

Following~\cite{dagan2024getting}, we defined compression using two metrics. 
The first, normalized sequence length (NSL), evaluates the efficiency of a tokenizer compared to our baseline \llama tokenizer. 
Formally, NSL $c_{\frac{\lambda}{\beta}}$ is defined as the ratio of the encoded sequence lengths from two tokenizers, $T_\lambda$ and $T_\beta$, across $N$ samples from dataset $D$.

$$
    c_{\frac{\lambda}{\beta}} = \frac{\sum_{i=1}^N |T_\lambda(D_i)|}{\sum_{i=1}^N |T_\beta(D_i)|}
$$

Just like the methodology in~\cite{dagan2024getting}, we employ the \llama tokenizer \cite{touvron2023llama} as our reference tokenizer $T_\beta$\footnote{This means that if $T_\lambda$ achieves an average NSL of $0.75$, it indicates that sequences encoded by $T_\lambda$ are 25\% shorter in terms of token count compared to those encoded by \llama.}.

\begin{table*}[th]
    \centering
\resizebox{0.8\textwidth}{!}{
\begin{tabular}{c c c c}
\toprule
\textsc{Approach} & \textsc{Normalized Sequence Length (NSL) ($\downarrow$)} & \textsc{Time Taken (hrs)}  \\ 
\midrule
\textsc{Multi-granular} & 0.58 & 30.13  \\
\textsc{Multiword-only} & 0.21 & 27.64  \\
\textsc{Target-only} & 0.32 & 27.20  \\
\textsc{Base-only} & 0.75 & 27.10   \\
\textsc{Merge} & 0.44 & 29.00  \\
\midrule
\bottomrule
\end{tabular}}
    \caption{\small Statistics of the selection process and the selected documents. \textsc{AVG. Sequence Length} is computed on randomly sampled 1000 documents from the pretraining set.}
    \label{tab:data_stats}
\end{table*}

\subsection{Comparison of Vocabulary Merging Techniques}\label{appendix:merger_2}

In terms of vocabulary merging, we also experiment with fixing the proportion of each type of token (\emph{subword}, \emph{word}, and \emph{multi-word}) in $v(t)$ at percentages \( p_{\text{subword}} = 0.6\), \(p_{\text{word}}  = 0.1\), and \(p_{\text{multi-word}} = 0.3 \), which we found to be the most performant combination.
However, fixed ratios do not work as well as the optimized vocabulary with vocabulary utility metric as described in the paper.

We also compare the proposed \textbf{multi-granular} sampling with different techniques of merging the target vocabulary set $V_{task}$ to the Llama-3 tokenizer $V_{start}$.
Several techniques are compared:
(1) \textbf{Merge}: we take the union of $V_{task}$ and $V_{start}$, but removing the duplicate tokens.
(2) \textbf{Target-only}: we use the task vocabulary set with the subword tokens.
(3) \textbf{Base-only}: we use the llama-3's vocabulary set with the subword tokens.
(4) \textbf{Multiword-only}: we use the acquired multi-word vocabulary that consists of only tokens concatenated by more than one word.
We show the results in Table~\ref{tab:merging_comparison}.

\begin{table*}[th]
    \centering
\resizebox{0.95\textwidth}{!}{
\begin{tabular}{c c c c c c c c c c}
\toprule
\multirow{2}{*}{\textsc{Approach}} & \multicolumn{9}{c}{\textsc{}} \\ 
& \textsc{ARC-Easy} & \textsc{ARC-Hard} & \textsc{BOOLQ} & \textsc{PIQA} & \textsc{SIQA}  & \textsc{HELLASWAG} & \textsc{OBQA} & \textsc{WINOGRANDE} & \textsc{Avg.} \\ 
\midrule
\multicolumn{9}{c}{125M params} \\ \midrule
\textsc{Multi-granular} & 44.24 & 30.13 & 57.98 & 72.71 & 41.16 & 46.68 & 35.74 & 52.10 & \textbf{47.59} \\
\textsc{Multiword-only} & 45.74 & 27.64 & 59.38 & 66.41 & 41.02 & 37.13 & 34.77 & 52.64 & 45.34 \\
\textsc{Target-only} & 43.59 & 27.20 & 57.76 & 72.17 & 42.29 & 45.89 & 31.05 & 50.24 & 46.14 \\
\textsc{Base-only} & 43.50 & 27.10 & 57.66 & 72.07 & 42.19 & 45.79 & 30.95 & 50.14 & 46.05 \\
\textsc{Merge} & 44.00 & 29.00 & 57.50 & 72.00 & 41.00 & 46.00 & 35.00 & 51.50 & 46.75 \\
\midrule
\bottomrule
\end{tabular}}
    \caption{\small Results over all downstream tasks based on data sampled with different granular features, each in terms of its respective metric on different granular tokens with a pretrained 125M model.}
    \label{tab:merging_comparison}
\end{table*}

\newpage

\subsection{Additional Results of the Impact of Domain Biases (1/2)
}\label{appendix:domain_bias_1}

Here we present the results as an extension for Figure~\ref{fig:domain}, where we present the results for selected data based on the rest of the seven benchmarks.
We show the results of multi-granular n-grams with n-gram baseline across eight tasks using 125M models.

\begin{figure}[th]
    \centering
    \begin{subfigure}[b]{\columnwidth}
        \centering
        \includegraphics[width=0.6\columnwidth]{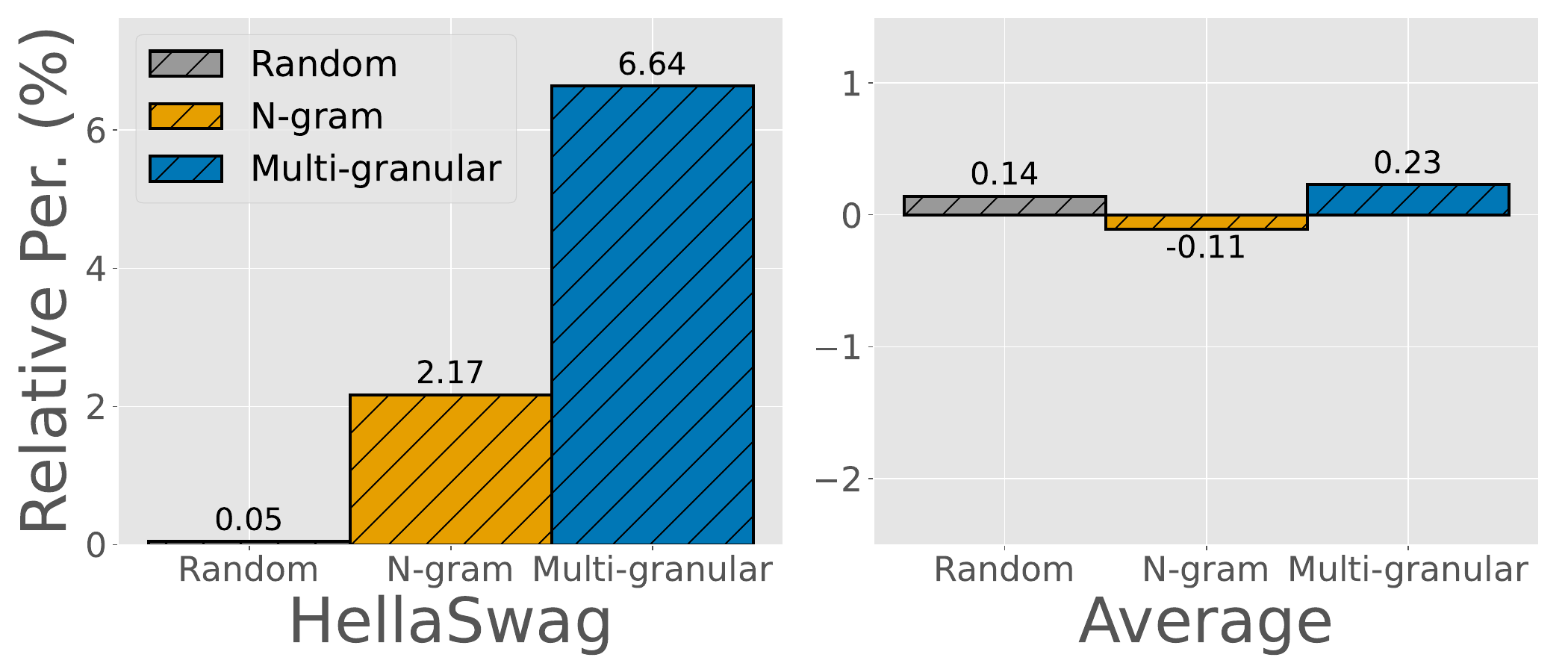}
        \label{fig:hellaswag}
    \end{subfigure}
    \vspace{10pt}  %
    \begin{subfigure}[b]{\columnwidth}
        \centering
        \includegraphics[width=0.6\columnwidth]{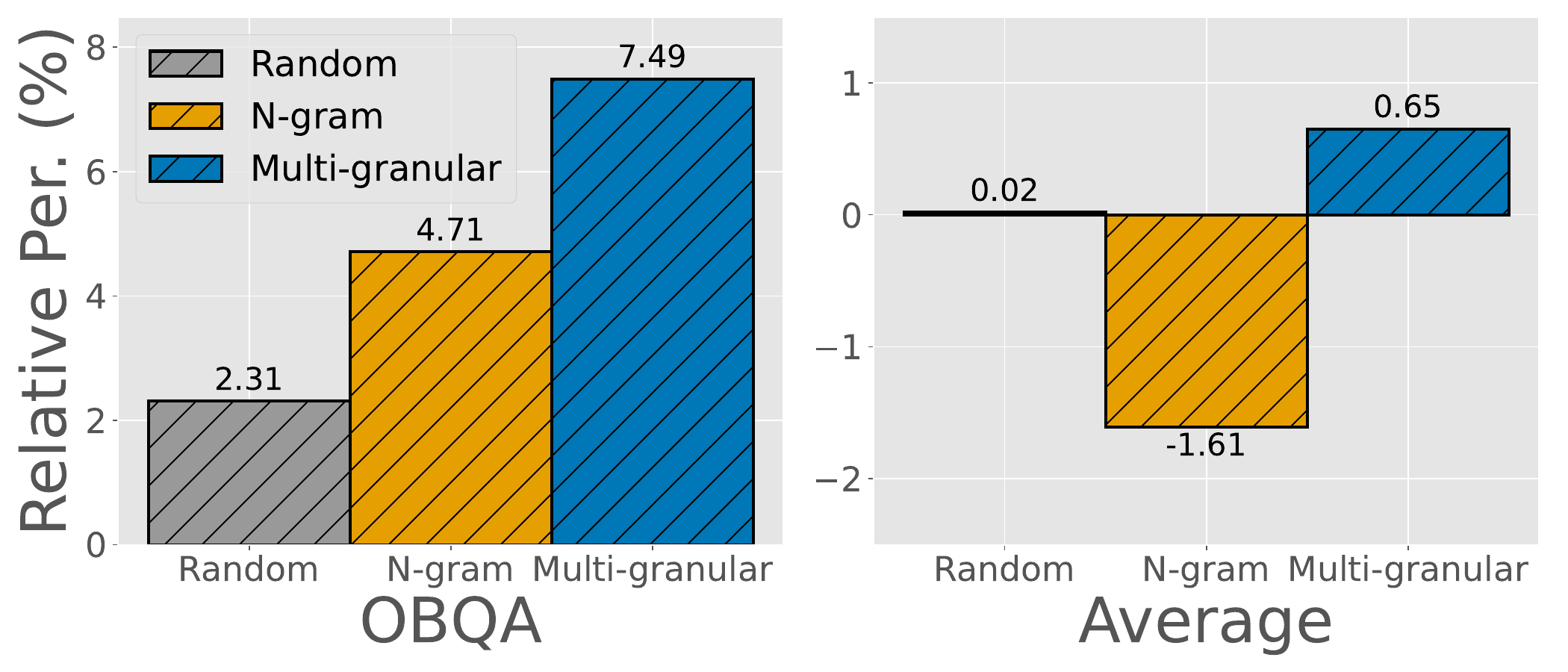}
        \label{fig:obqa}
    \end{subfigure}
    \vspace{10pt}  %
    \begin{subfigure}[b]{\columnwidth}
        \centering
        \includegraphics[width=0.6\columnwidth]{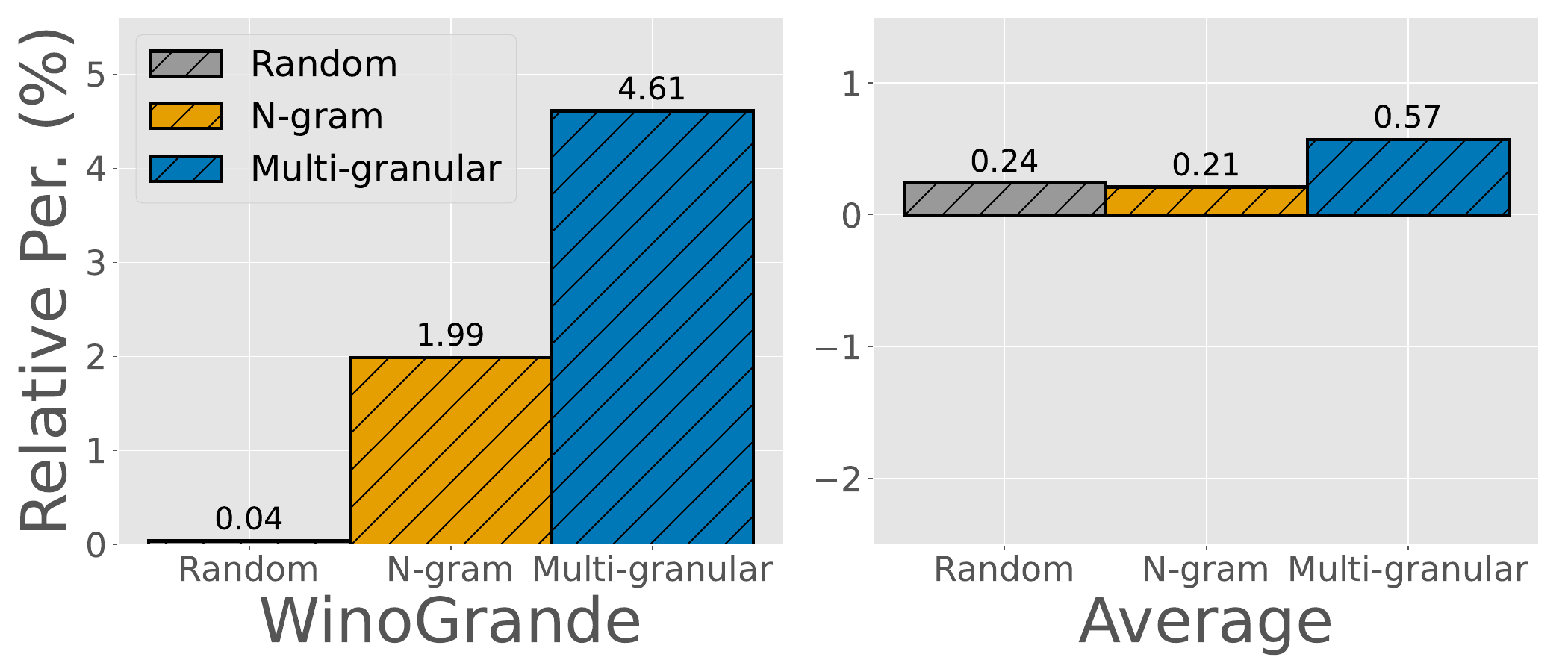}
        \label{fig:winogrande}
    \end{subfigure}
\caption{\small 
Comparison of \textbf{Multi-granular} n-grams with \textbf{N-gram} and \textbf{Random} baseline using 125M models, trained solely on data selected with HellaSwag, OBQA, WinoGrande
data as the target respectively.}
\end{figure}

\newpage
\subsection{Additional Results of the Impact of Domain Biases (2/2)
}\label{appendix:domain_bias_2}

\begin{figure}[ht]
    \centering
    \begin{subfigure}[b]{\columnwidth}
        \centering
        \includegraphics[width=0.6\textwidth]{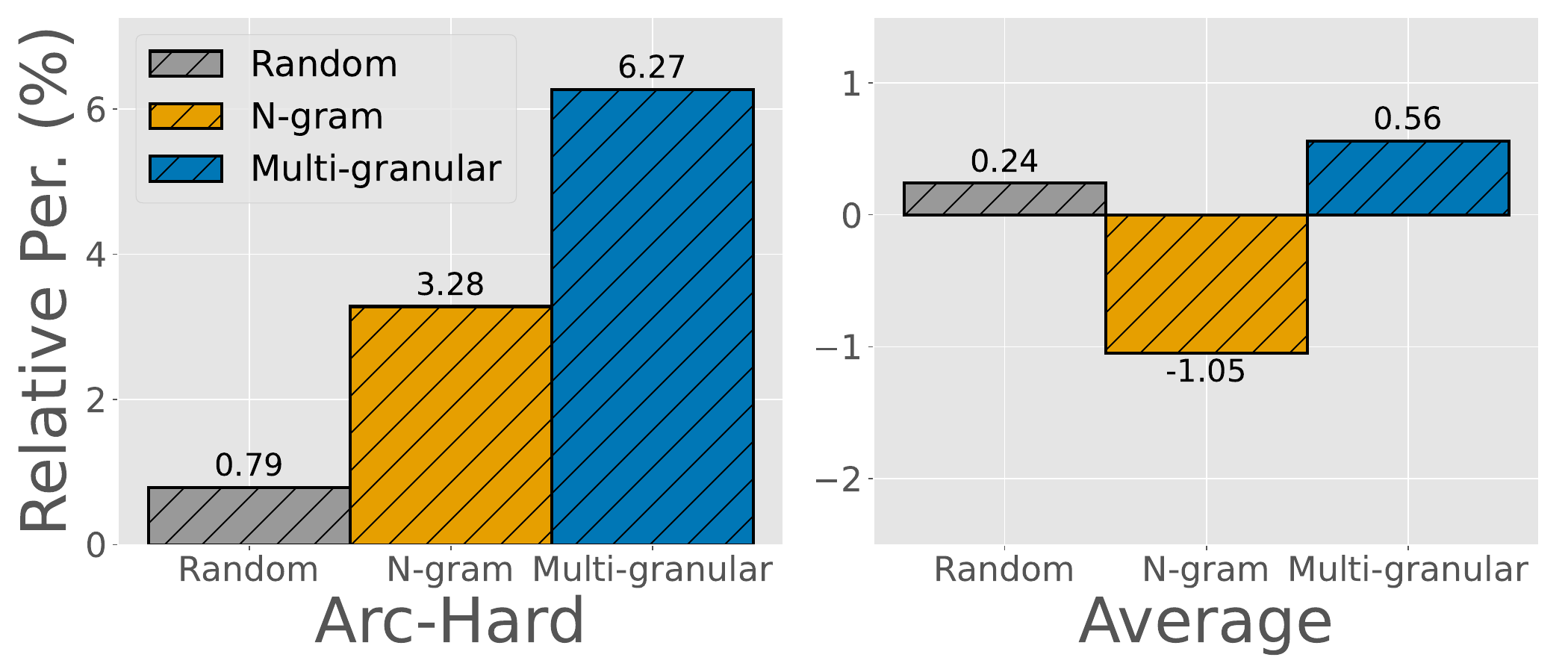}
        \label{fig:arc-hard}
    \end{subfigure}
    \vspace{2pt}  %
    \begin{subfigure}[b]{\textwidth}
        \centering
        \includegraphics[width=0.6\textwidth]{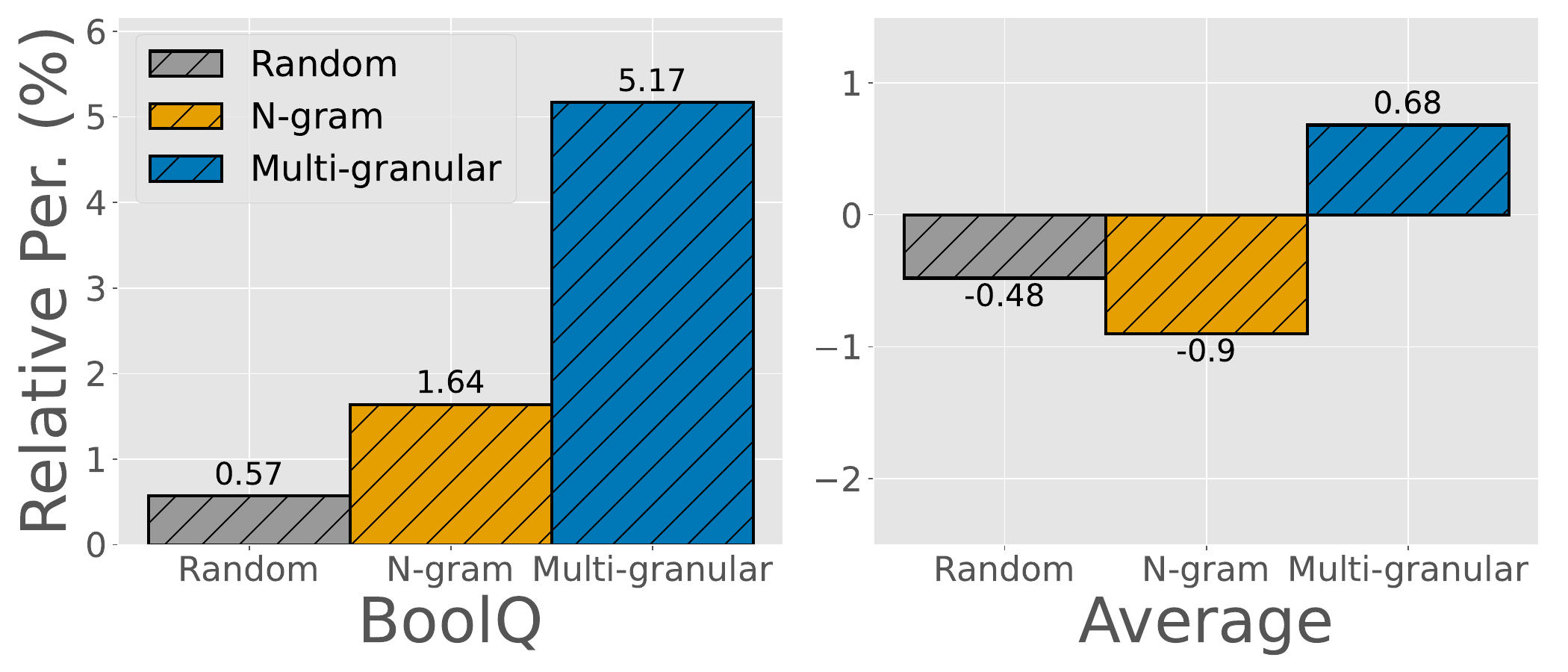}
        \label{fig:boolq}
    \end{subfigure}
    \begin{subfigure}[b]{\textwidth}
        \centering
    \includegraphics[width=0.6\textwidth]{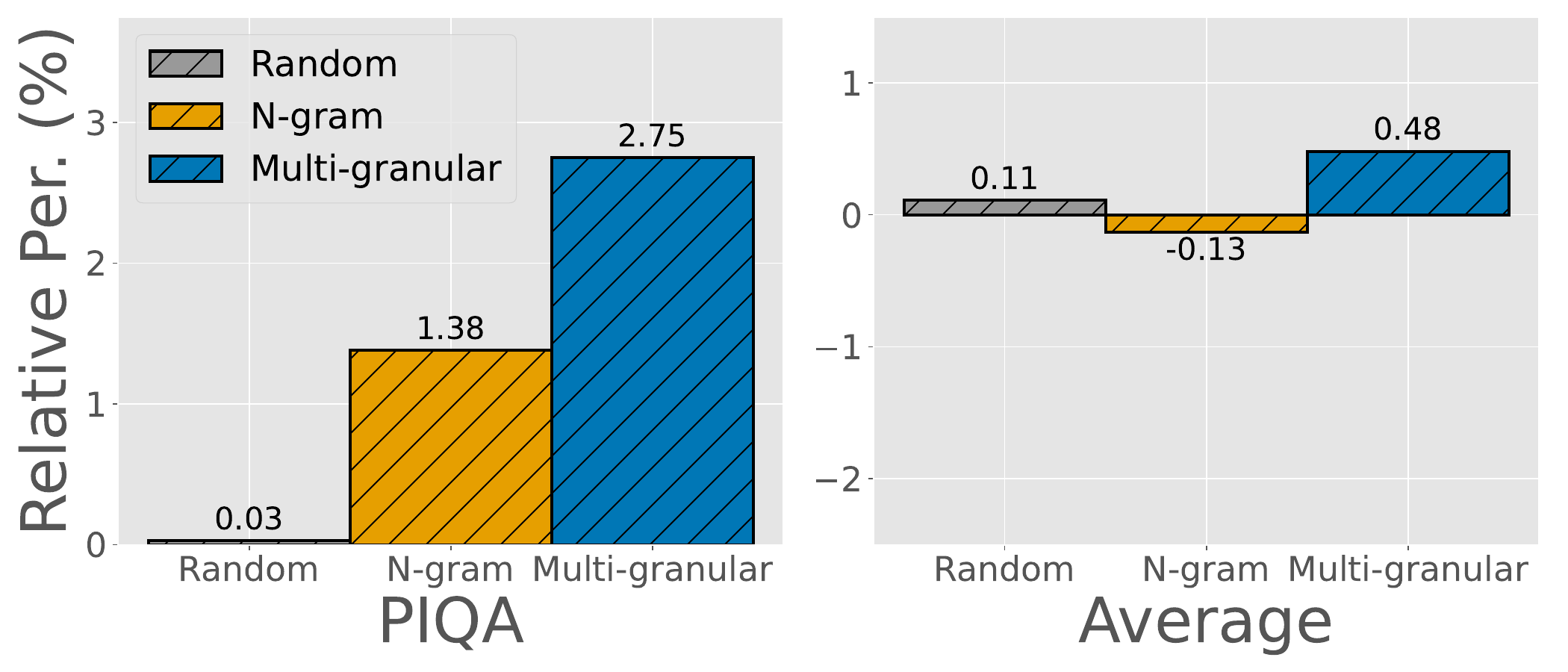}
        \label{fig:piqa}
    \end{subfigure}
    \vspace{2pt}  %
    \begin{subfigure}[b]{\textwidth}
        \centering
        \includegraphics[width=0.6\textwidth]{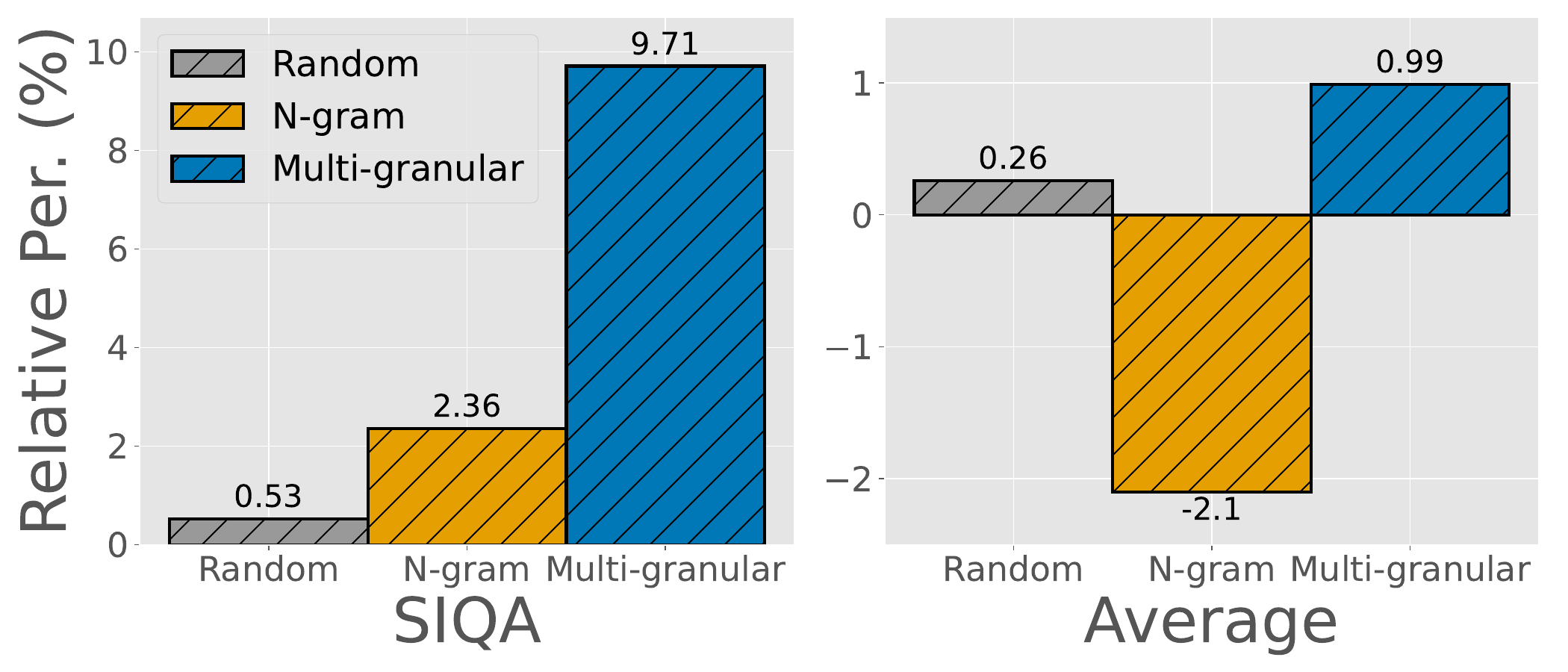}
        \label{fig:siqa}
    \end{subfigure}
    \caption{\small 
Comparison of \textbf{Multi-granular} n-grams with \textbf{N-gram} and \textbf{Random} baseline using 125M models, trained solely on data selected with Arc-Hard, BoolQ, PIQA, and SIQA data as the target respectively.}
\end{figure}